\documentclass[lettersize,journal]{IEEEtran}
\usepackage{amsmath,amsfonts}
\usepackage{algorithmic}
\usepackage{algorithm}
\usepackage{array}
\usepackage[caption=false,font=normalsize,labelfont=sf,textfont=sf]{subfig}
\usepackage{textcomp}
\usepackage{stfloats}
\usepackage{verbatim}
\usepackage{graphicx}
\usepackage{gensymb}
\usepackage{cite}
\usepackage{multirow}
\usepackage{booktabs}
\usepackage{colortbl}
\usepackage{todonotes}

\definecolor{oddrowcolor}{RGB}{220, 220, 220}  
\definecolor{evenrowcolor}{RGB}{255, 255, 255} 

\usepackage[hyphens,spaces,obeyspaces]{url}
\usepackage[colorlinks,allcolors=blue]{hyperref}

\usepackage[nolist,nohyperlinks]{acronym}

\acrodef{us}[US]{ultrasound}
\acrodef{cnn}[CNN]{convolutional neural network}
\acrodef{mse}[MSE]{mean square error}
\acrodef{ct}[CT]{computerised tomography}
\acrodef{mri}[MRI]{magnetic resonance imaging}
\acrodef{ga}[GA]{gestational age}
\acrodef{dl}[DL]{deep learning}
\acrodef{ml}[ML]{machine learning}
\acrodef{rl}[RL]{reinforcement learning}
\acrodef{drl}[DRL]{deep reinforcement learning}
\acrodef{irl}[IRL]{inverse reinforcement learning}
\acrodef{dof}[DoF]{degree of freedom}
\acrodef{6d}[6D]{six-dimensional}
\acrodef{4d}[4D]{four-dimensional}
\acrodef{3d}[3D]{three-dimensional}
\acrodef{2d}[2D]{two-dimensional}
\acrodef{sp}[SP]{standard plane}
\acrodef{nhs}[NHS]{National Healthcare System}
\acrodef{ai}[AI]{Artificial Intelligence}
\acrodef{csp}[CSP]{cavum septum pellucidum}
\acrodef{lv}[LV]{lateral ventricle}
\acrodef{tt}[TT]{transthalamic}
\acrodef{tv}[TV]{transventricular}
\acrodef{tc}[TC]{transcerebellar}
\acrodef{rnn}[RNN]{recurrent neural network}
\acrodef{sl}[SL]{supervised learning}
\acrodef{marl}[MARL]{Multi-Agent Reinforcement Learning}
\acrodef{hc}[HC]{head circumference}
\acrodef{bmi}[BMI]{body mass index}
\acrodef{sac}[SAC]{Soft Actor Critic}
\acrodef{ppo}[PPO]{Proximal Policy Optimization}
\acrodef{td3}[TD3]{Twin Delayed Deep Deterministic Policy Gradient}
\acrodef{dqn}[DQN]{deep Q-learning Network}
\acrodef{lfd}[LfD]{Learning from Demonstration}
\acrodef{loocv}[LOOCV]{Leave One Out Cross-Validation}
\acrodef{mdp}[MDP]{Markov Decision Process}
\acrodef{dicom}[DICOM]{Digital Imaging and Communications in Medicine}
\acrodef{roi}[ROI]{Region of Interest}
\acrodef{mmi}[MMI]{Mattes Mutual Information}
\acrodef{rms}[RMS]{Root Mean Square}
\acrodef{dsr}[DSR]{Direct Simultaneous Registration}

\usepackage{eso-pic}

\AddToShipoutPictureBG*{%
   \AtPageUpperLeft{\put(60,-6){\scriptsize This article has been accepted for publication in IEEE Transactions on Medical Robotics and Bionics. This is the author's version which has not been fully edited and}}
   \AtPageUpperLeft{\put(150,-15){\scriptsize content may change prior to final publication. Citation information: DOI 10.1109/TMRB.2023.3328638.}}
   \AtPageLowerLeft{\put(115,10){\scriptsize This work is licensed under a Creative Commons Attribution 4.0 License. For more information, see \url{https://creativecommons.org/licenses/by/4.0/}}}
}

\begin{document}

\title{Ultrasound plane pose regression: assessing generalized pose coordinates in the fetal brain}

\author{Chiara~Di~Vece,~\IEEEmembership{Student Member,~IEEE}, Maela~Le~Lous, Brian~Dromey, Francisco~Vasconcelos, Anna~L~David, Donald~Peebles, Danail~Stoyanov~\IEEEmembership{Senior Member,~IEEE}

\thanks{This research was funded in whole, or in part, by the Wellcome/EPSRC Centre for Interventional and Surgical Sciences (WEISS) [203145/Z/16/Z]; the Engineering and Physical Sciences Research Council (EPSRC) [EP/P027938/1, EP/R004080/1, EP/P012841/1]; the Royal Academy of Engineering Chair in Emerging Technologies Scheme, and Horizon 2020 FET Open (863146). For the purpose of open access, the author has applied a Creative Commons Attribution (CC BY) licence to any Author Accepted Manuscript version arising.}
\thanks{C. Di Vece, F. Vasconcelos and D. Stoyanov are with Wellcome/EPSRC center for Interventional and Surgical Sciences (WEISS) and the Department of Computer Science, University College London (UCL), UK. M. Le Lous, B. Dromey, A.L. David and D.Peebles are with WEISS, Elizabeth Garrett Anderson Institute for Women's Health and NIHR University College London Hospitals Biomedical Research center, UCL, UK.\\chiara.divece.20@ucl.ac.uk}}

\markboth{Journal of \LaTeX\ Class Files,~Vol.~14, No.~8, August~2021}%
{Di Vece \MakeLowercase{\textit{et al.}}: Learning\ac{us} plane pose regression: assessing generalized pose coordinates in the fetal brain}

\IEEEpubid{0000--0000/00\$00.00~\copyright~2021 IEEE}

\maketitle

\begin{abstract}
In obstetric \ac{us} scanning, the learner's ability to mentally build a \ac{3d} map of the fetus from a \ac{2d} \ac{us} image represents a significant challenge in skill acquisition. We aim to build a \ac{us} plane localization system for \ac{3d} visualization, training, and guidance without integrating additional sensors. This work builds on top of our previous work, which predicts the \ac{6d} pose of arbitrarily oriented \ac{us} planes slicing the fetal brain with respect to a normalized reference frame using a \ac{cnn} regression network. Here, we analyze in detail the assumptions of the normalized fetal brain reference frame and quantify its accuracy with respect to the acquisition of \ac{tv} \ac{sp} for fetal biometry. We investigate the impact of registration quality in the training and testing data and its subsequent effect on trained models. Finally, we introduce data augmentations and larger training sets that improve the results of our previous work, achieving median errors of 2.97 $mm$ and 6.63\degree~for translation and rotation, respectively.
\end{abstract}


\section{Introduction}
\label{sec:intro}

\begin{figure}[!t]
    \centering
    \includegraphics[width=\columnwidth]{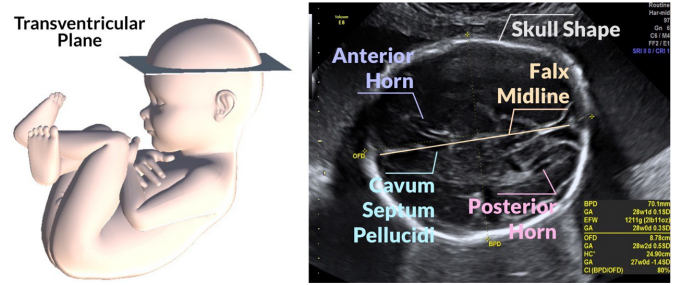} 
    \caption{TV SP to evaluate fetal biometry in the brain. This plane needs to show the skull shape, the CSP, the posterior horn of the lower lateral ventricle, the anterior horns of the lateral ventricles, the skull shape, and the falx midline}
    \label{fig:sp}
\end{figure}

\IEEEPARstart{F}{etal} \ac{us} is a non-invasive, real-time, and cost-effective diagnostic tool for monitoring fetal growth and anatomy throughout gestation~\cite{Hadlock1985}. During routine mid-trimester fetal \ac{us} scan, the sonographer acquires the \acs{sp}, predefined anatomical planes defined by scientific committees to promote international guidelines for fetal \ac{us} images~\cite{Salomon2011}. Specifically, the \ac{tv} \ac{sp} needs to show the skull shape, the \ac{csp}, the posterior horn of the lower lateral ventricle, and the anterior horns of the lateral ventricles~\cite{Salomon2013} (Figure~\ref{fig:sp}). This allows for reliable measurements of specific structures and reduced inter- and intra-sonographer variability. The correct identification of \acp{sp} is essential in the second-trimester fetal anatomic survey to investigate the morphological characteristics of the fetus and detect abnormalities or deviations from the expected growth patterns. Sonographers may struggle to obtain good \acp{sp} for various reasons, including inexperience, limited training, time limitations, and fetal movement~\cite{Sarris2011,Bahner2016}. Most trainees learn to scan actual patients under the direct supervision of an expert. Although \ac{us} simulators have been developed recently, trainee engagement has been limited due to competing time priorities~\cite{Chandrasekaran2016SimulationPDF}. The primary training challenge faced by all novice sonographers is not related to knowledge of anatomy or familiarity with the \ac{us} machine interface. Rather, the manual navigation of the probe toward acquiring \ac{sp} requires the sonographer to build a \ac{3d} map of the fetus from dynamic \ac{2d} sectional views while handling the probe. Measurements of biometric parameters and assessments of the fetal brain's anatomy may be erroneous due to mistakes in locating the \ac{2d} scan within the \ac{3d} volume.
\IEEEpubidadjcol

At present, \acp{sp} recognition represents the main focus of fetal \ac{us} training. Due to the requirement to interpret variable and complex images and their spatial relationship, autonomous probe navigation toward a target plane remains challenging~\cite{Li2021AnAcquisitions}. Our final aim is to develop a \ac{us} navigation system that guides the sonographer toward obtaining \acp{sp} with reference to fetal anatomy.

 In~\cite{DiVece2022}, we proposed a method to localize a \ac{us} plane of a fetal brain directly from its \ac{2d} \ac{us} image. While this is a promising result toward active guidance during fetal scanning, a few aspects still require further investigation. First, it assumes that brains from different fetuses can be mapped to the same normalized coordinate system, where each \ac{sp} always has the same pose coordinates. The accuracy of this assumption has not been quantified, which limits the analysis of experimental results. Secondly, the achieved pose accuracy is still far from optimal due to a lack of variation in training data.

In this work, we expand on this work with the following contributions:
\begin{itemize}
    \item We developed a tool for annotating poses of \acp{sp} in \ac{3d} \ac{us} volumes using the Unity engine and a gaming controller\footnote{\color{blue}\href{https://github.com/surgical-vision/FetalUltrasoundSimulator-Unity.git}{FetalUltrasoundSimulator-Unity - GitHub}}. We used this tool to obtain ground truth poses of \ac{tv} \acp{sp} (annotated by an obstetrician) of a publicly available fetal brain dataset. This provides a novel way of assessing the registration of fetal brain volumes from different fetuses.
    \item We evaluate to which extent the brains of different fetuses can be registered to the same generalized coordinate frame. While this is essential to assess training and validation data for fetal brain pose regression models, it has not been done in the previous literature on this topic~\cite{DiVece2022,Yeung2021,Yeung2022}. 
    \item We evaluate different volume registration techniques and observe that manually annotating anatomical fiducials within the fetal brain is fundamental for good registration results. We demonstrate that this also contributes to the effective training of pose regression models and preserving their assumptions. 
    \item We outperform the pose regression results obtained in~\cite{DiVece2022} by improving the volume registration procedure, introducing additional data augmentation, and increasing training sets.
\end{itemize}

\section{Related Work}
\label{sec:related}

This section presents the related work for three related but different tasks: extraction of \acp{sp}, slice-to-volume registration, and localization of \acp{sp}.

\subsection{Extraction of Standard Planes}
\label{subsec:related-sps}

Previous work proposed automating the extraction of \acp{sp} from data acquired with a simplified protocol rather than assisting operators in acquiring typical freehand \ac{2d} \acp{sp}. In one of their initial works, Zhang \emph{et al.}~\cite{Zhang2012} developed a system based on two AdaBoost classifiers placed in cascade to automatically detect in a coarse-to-fine way early gestational sac \ac{sp}. Further early studies~\cite{Ni2013,Ni2014,Yang2014,Lei2014} detect key abdominal structures and landmarks in a sequence of \ac{2d} \ac{us} fetal images to classify the \acp{sp} in each frame of \ac{us} videos based on the presence and orientation of the landmarks using various conventional \ac{ml} algorithms like AdaBoost, Random Forest as well as support vector machines. These methods, however, are only applicable to a subset of fetal \acp{sp} (brain and abdomen); besides, the quality of the obtained \ac{sp} cannot be compared with the one achieved with typical freehand scanning.

Classification methods based on \acp{cnn} were used to detect \ac{2d} \acp{sp} because of their powerful ability to learn hierarchical representations automatically. To detect the fetal \acp{sp}, Chen \emph{et al.}~\cite{Chen2015} fine-tuned a pre-trained classification \ac{cnn} based on transfer learning. Baumgartner \emph{et al.}~\cite{Baumgartner2017a} proposed a classification model to detect thirteen \acp{sp} with unsupervised learning and then used weakly-\ac{sl} based on image-level labels to locate anatomical structures in each plane. The study employed extensive data, including videos longer than those usually collected in clinical practice (roughly 30 minutes). The \acp{cnn} are fed with surrounding and additional information from each \ac{us} video. To capture temporal information in \ac{2d} \ac{us}, some works added to the detection of the three fetal \acp{sp} a \ac{rnn}~\cite{Lin2019a}.

All the methods mentioned above are effective in the detection of \ac{sp} images. Still, they can only determine whether an image was captured at a \ac{sp}, not where exactly it is in the corresponding \ac{3d} space. Besides, the models require a high amount of annotated data to be trained.

\subsection{Slice-to-volume Registration}
\label{subsec:related-s2vreg}

One approach for \ac{us} plane localization is to find its alignment with respect to a pre-acquired \ac{3d} volume of the same anatomy. This optimization problem is typically solved with iterative numerical methods that minimize the distance between specific landmarks or maximize intensity-based similarity metrics~\cite{Kitchen1994}. Unfortunately, the cost functions associated with these metrics are frequently non-convex, limiting the capture range of these registration methods. Our task differs from a classic slice-to-volume registration method, \emph{i.e.}, it does not require a previously acquired \ac{3d} volume of the same subject being scanned. Instead, we predict the pose relative to a generalized brain center, \emph{i.e.}, a stable anatomical brain point across the different, pre-aligned volumes, where training and test data belong to different subjects.

\subsection{Localization of Standard Planes}
\label{subsec:related-localization}

Predicting the pose of \acp{sp} in \ac{3d} volumes can be performed without a patient-specific model and without using pre-operative data. Various methods have been proposed for the localization of \ac{us} planes and \ac{us} probe navigation using reinforcement learning~\cite{Li2021AutonomousLearning,Bi2022VesNet-RL:Navigation} or external sensors~\cite{Droste2020a,birlo2022}. In the context of fetal scanning, this has been primarily approached as a classification problem, where the plane pose space is discretized into bins, and the estimation boils down to selecting one of the bins~\cite{Tulsiani2015,Su2015}. In fetal \ac{mri}~\cite{Hou2017,Hou2018} and fetal \ac{us}~\cite{Namburete2018}, the prediction of slice locations has been previously improved with learning-based methods. General purpose learning-based methods for pose estimation approach this as a regression of a \ac{3d} translation and a \ac{3d} rotation. \ac{3d} rotations can be represented in conventional ways, such as quaternions, axis-angle, or Euler angles. Zhou et al. pointed out in~\cite{Zhou2019} that if the entire rotation space is required, these representations are sub-optimal for specific angle ranges, and proposed a new \ac{6d} representation for rotations that does not suffer from these issues. This rotation representation has been adopted for \ac{us} plane localization in~\cite{DiVece2022}, where a regression \ac{cnn} is proposed to predict the \ac{6d} pose of arbitrarily-oriented planes slicing the fetal brain \ac{us} volume without the need for real ground truth data in real-time or \ac{3d} volume scans of the patient beforehand. The proposed network reliably localizes \ac{us} planes within the fetal brain in phantom data and successfully generalizes pose regression for unseen fetal brains from a similar \ac{ga} as in training. The network was tested on real fetal brain images with a \ac{ga} ranging from 21 to 25 weeks. Similarly, Yeung et al.~\cite{Yeung2021} proposed a \ac{cnn} that takes a set of images as input and learns to compare them in pairs. The model was tested on fetal brain volumes with a \ac{ga} ranging from 18 to 22 weeks. Then, in~\cite{Yeung2022}, an unsupervised cycle consistency using the fact that the overall displacement of a sequence of images in the \ac{3d} anatomical atlas is equal to the displacement from the first image to the last in that sequence was added. 

\section{Materials and Methods}
\label{sec:methods}

\begin{figure*}[!t]
    \centering
    \includegraphics[width=\textwidth]{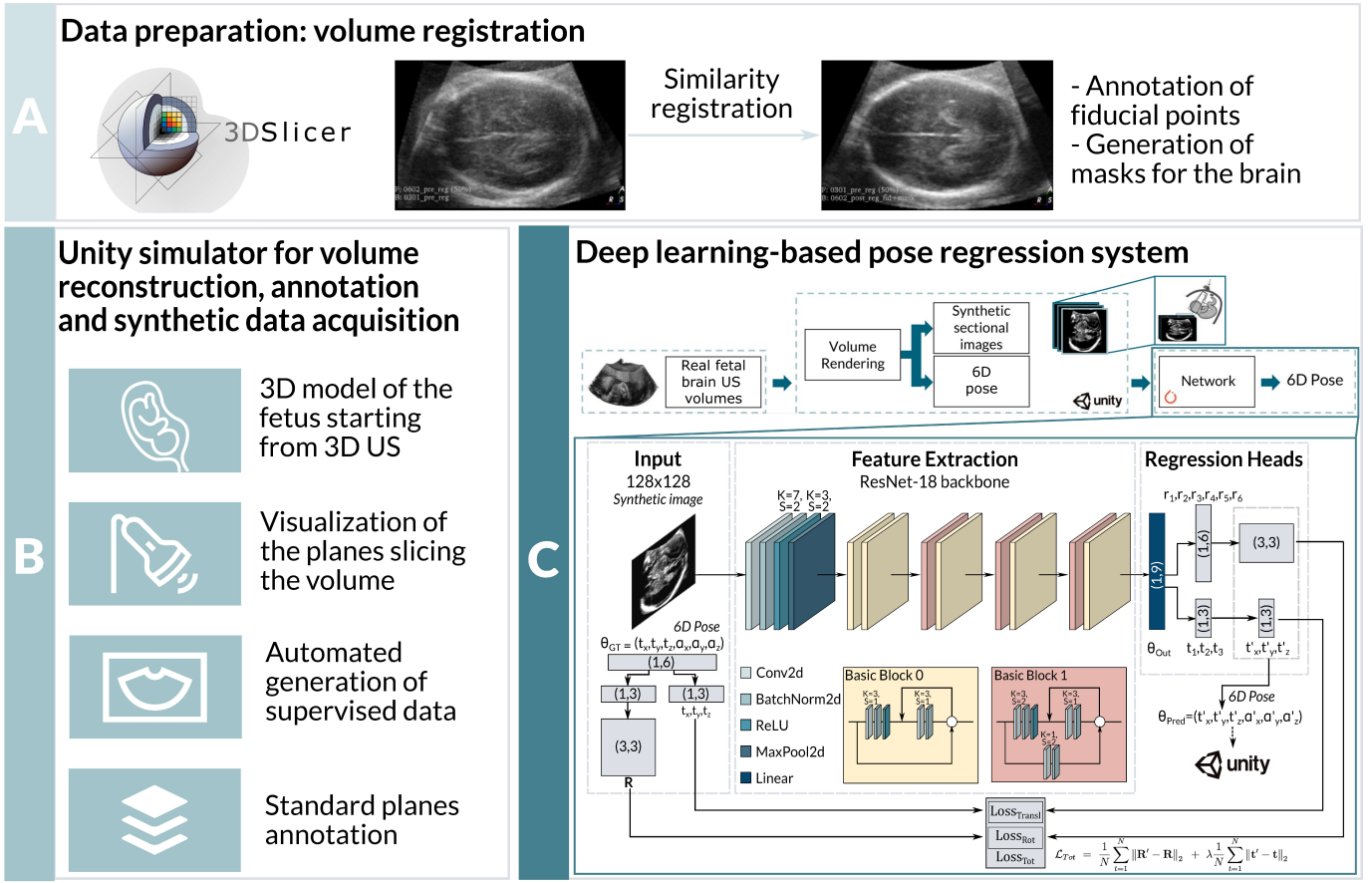}    
    \caption{Pipeline to train and test the network. (A) The \ac{3d} fetal brain \ac{us} volumes are registered in \ac{3d} Slicer using similarity registration; (B) The volumes are reconstructed into Unity, and synthetic sectional (slice) image representations are generated and saved along with their 6D pose (translation and rotation) relative to the center of the fetal brain \ac{us} volume; (C) These images are fed into the network to output the estimated slice 6D pose (translation and rotation) relative to the same point}
    \label{fig:pipeline}
\end{figure*}

The development of our \ac{us} pose regression system has been divided into three main blocks, reported in Figure~\ref{fig:pipeline}. First, we align \ac{3d} \ac{us} volumes for the training and validation of our models. Secondly, we developed a Unity-based simulator to visualize and manually annotate \acp{sp} in \ac{3d} \ac{us} volumes. This also enables the automated generation of supervised training data for our pose regression models, \emph{i.e.}, \ac{2d} synthetic images, and their ground truth \ac{6d} pose relative to the volume center. Finally, we detail our \ac{dl}-based plane pose regression system.
\subsection{Data preparation}
\label{subsec:methods-data}

\subsubsection{Dataset}
\label{subsubsec:methods-data-dataset}

We utilize a selection of 6 fetal brain \ac{us} volumes from a dataset of 188 volumes~\cite{Pistorius2010} (singleton pregnancy with no abnormal findings)\footnote{Refer to {\color{blue}\href{https://dataverse.nl/dataset.xhtml?persistentId=doi:10.34894/X0Z7U1}{www.datavers.nl}} for details}. The criteria for our selection is to test the generalization of a canonical frame to different fetuses within a 20-25 weeks range, where key anatomical landmarks can be clearly annotated for registration and evaluation. In detail, we select volumes acquired at the axial \ac{tv} \ac{sp} position (166 volumes); we only include volumes within the 20-25 weeks \ac{ga} range (46 volumes); lastly, an experienced obstetrician excluded volumes where key landmarks (such as optic nerve) were not clearly visible along with multiple scans of the same fetus. This resulted in six real fetal brain \ac{us} volumes with a \ac{ga} ranging from 21 to 25 obtained from different fetuses ($f_{i}$, with $i=1,...,6$). All volumes were processed to be isotropic with a voxel size of 0.5$\times$0.5$\times$0.5~$mm$ and an average size of 249$\times$174$\times$155~$mm$ (\emph{coronal}$\times$\emph{axial}$\times$\emph{sagittal}, actual size of the acquired volumes).

\subsubsection{Volume registration}
\label{subsubsec:methods-data-reg}

\begin{figure*}[!t]
     \centering
     \includegraphics[width=\textwidth]{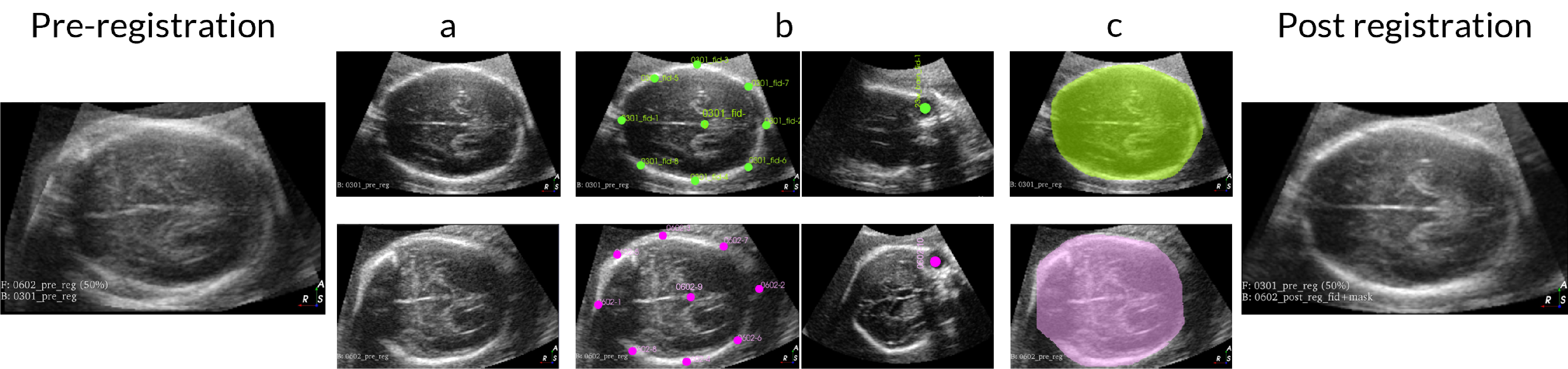}
     \caption[Registration procedure on real fetal brain \ac{us} volumes]{Registration procedure on real fetal brain \ac{us} volumes. Workflow from pre-registration to post-registration: (a) Starting \ac{us} fetal brain volumes before registration, (b) Some of the fiducial points used to achieve an initial alignment of the volumes, (c) Contour mask of the brain used to avoid overfitting on the shape of the \ac{us} volume}
     \label{fig:reg}
 \end{figure*}

Even though all scans were acquired with a single protocol and in the position of the \ac{tv} \ac{sp}, experiments show that the anatomy within those volumes is far from well aligned and requires further registration.
We tested different registration methods and obtained the best results with a fiducial-based approach utilizing 3D Slicer~\cite{3dslicerweb,3dslicer}. The method is depicted in Figure~\ref{fig:reg}. Starting from the initial volumes (Figure~\ref{fig:reg}a), we proceed as follows:

\begin{itemize}
    \item We annotated fiducial points to achieve an initial alignment of the volumes using the Fiducial Registration Wizard module (Figure~\ref{fig:reg}b);
    \item Defined a contour mask of the brain using the Segment Editor module to avoid overfitting on the shape of the \ac{us} volume during registration;
    \item Used the general registration (BRAIN) module available in 3D Slicer to register the volumes with a similarity registration phase (rigid registration + scale for a total of 7 degrees of freedom), as shown in Figure~\ref{fig:reg}c. We chose the previously obtained masks as a \ac{roi} so that the registration algorithm only considers a specific image region for the registration (the fetal brain).
\end{itemize}
 
\subsection{Unity Simulator for Standard Plane Annotations and Synthetic Images Generation}
\label{subsec:methods-unity}

Starting from an open-source project\footnote{\textcolor{blue}{\href{https://github.com/mlavik1/UnityVolumeRendering/}{UnityVolumeRendering}}}, we developed our simulator using the game engine Unity~\cite{unity3d}.
The first step is rendering the volume starting from RAW, PARCHG, or \ac{dicom} datasets. The simulator allows the user to render the volume using three modes: Isosurface Rendering, Maximum Intensity Projection, and Direct Volume Rendering. The latter is the standard rendering mode; it uses transfer functions (1D or \ac{2d}) to determine the color and opacity while projecting rays across the dataset. Transfer functions translate density (and gradient magnitude in case of \ac{2d}) to a color and opacity. The simulator allows the user to set a custom transfer function.
\begin{itemize}
    \item 1D Transfer Function: the density is represented on the X-axis, whereas the opacity (alpha) is on the Y-axis. The user can create a curve for opacity by density by shifting the grey alpha knots. The bottom gradient-colored panel maps color to density.
    \item \ac{2d} transfer function: the density is represented on the X-axis, whereas the gradient magnitude is on the Y-axis. Using the sliders, the user can define a rectangle shape, modify its size/position, and the minimum and maximum values for alpha/opacity.
\end{itemize}

\subsubsection{User interface}
\label{subsubsec:methods-unity-ui}

\begin{figure}[!t]
     \centering
     \includegraphics[width=\columnwidth]{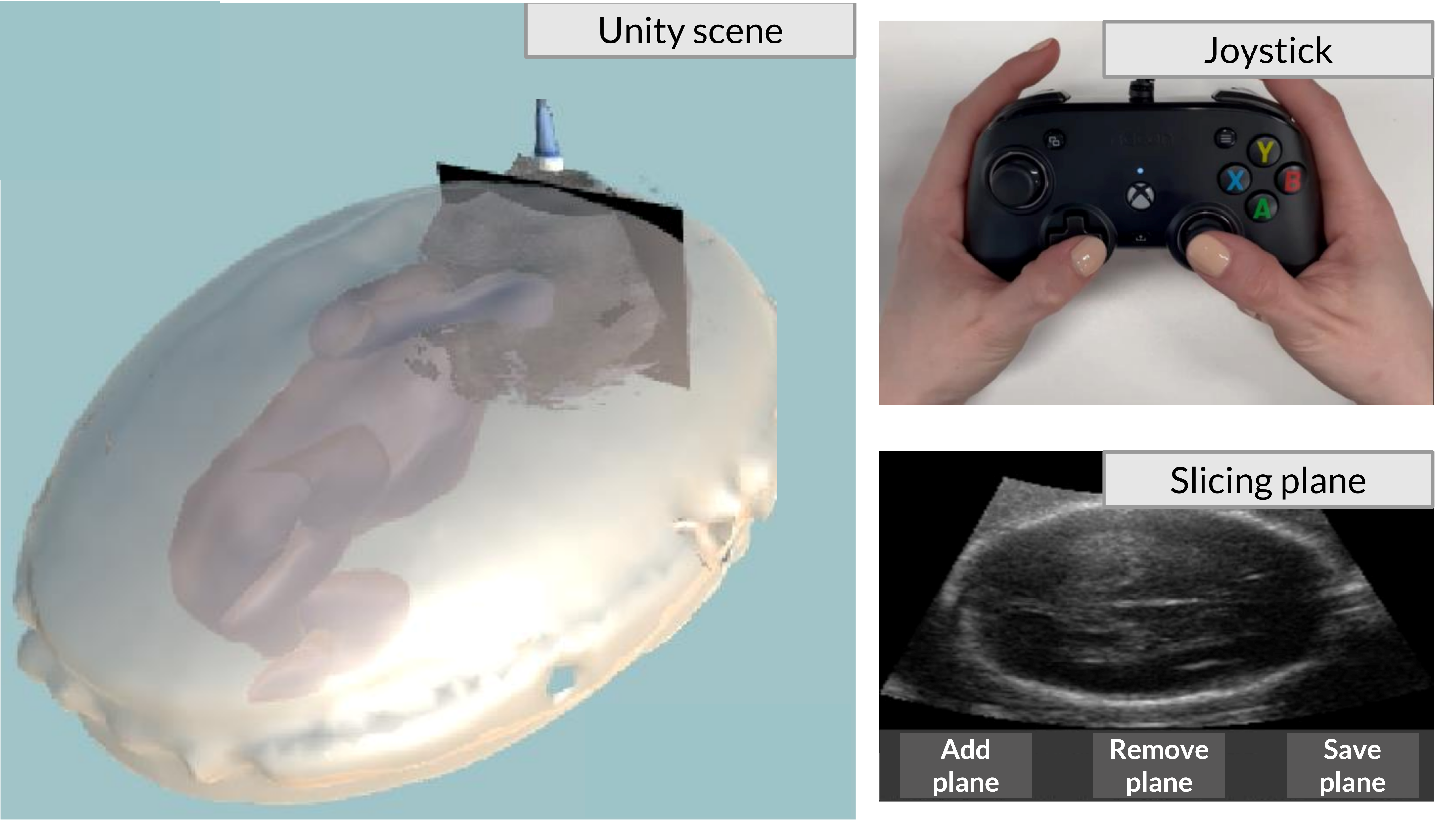}
     \caption[Unity simulator for volume reconstruction, SP annotations, and automatic supervised data generation]{Unity simulator for volume reconstruction, SP annotations, and automatic supervised data generation. Using the suggested commands, the clinician can visualize the slicing plane in an external window while controlling the probe with the joystick. Once the desired plane is reached, it can be saved using the ``Save plane'' button along with its 6D pose relative to the volume center}
     \label{fig:unity}
 \end{figure}

After loading the \ac{dicom} dataset extracted from \ac{3d} Slicer and setting the transfer function, the clinician can add a plane slicing the reconstructed \ac{3d} \ac{us} volume (left-hand side of Figure~\ref{fig:unity}) with an arbitrary orientation and visualize the plane in an external window (right-hand side, bottom). The plane can be controlled using a joystick (right-hand side, top), simplifying the annotation of the \acp{sp}. Hence, the clinician can modify the position and rotation of the plane using the joystick while monitoring the appearance in the external window. Once the clinicians are satisfied with the pose of the \ac{sp}, they can save it using the last button in the external windows to get a picture of the slicing plane and its \ac{6d} pose relative to the volume center. 
\subsubsection{Training and testing data generation}
\label{subsubsec:methods-unity-datagen}

To generate training data for our models, we generated synthetic slices by applying rotation and translation to a plane placed in the center of the volume generated with a uniform random distribution within a fixed range to avoid slices with poor overlap with the volume. The synthetic images obtained by slicing the volume were saved along with their pose with respect to the volume center (fetal brain). This provides an automated way of generating a high amount of training data with reliable ground truth labels. 
An obstetrician annotated the position of the \ac{tv} \ac{sp} by directly manipulating a slicing plane with the joystick and choosing the translation and angle sampling intervals to avoid sampling planes at the edges of the volume containing no information. The nearby planes were generated by applying small random rotations and translations (uniform distribution). Specifically, the acquisition interval between two planes was decreased from 0.1 to 0.001 for translation (Unity environment, with coordinates normalized between -1 and 1 so that the pose regression works in a fixed, normalized range, independent of the real brain size in $mm$) and from 7.9\degree~to 1.9\degree~for rotation compared to the acquisition of planes at random coordinates. We acquired 20699 planes with random orientation per volume and 1330 around the \ac{tv} \ac{sp} for a total of 22029 images for each volume.

  \begin{figure*}[!t]
     \centering
     \includegraphics[width=\textwidth]{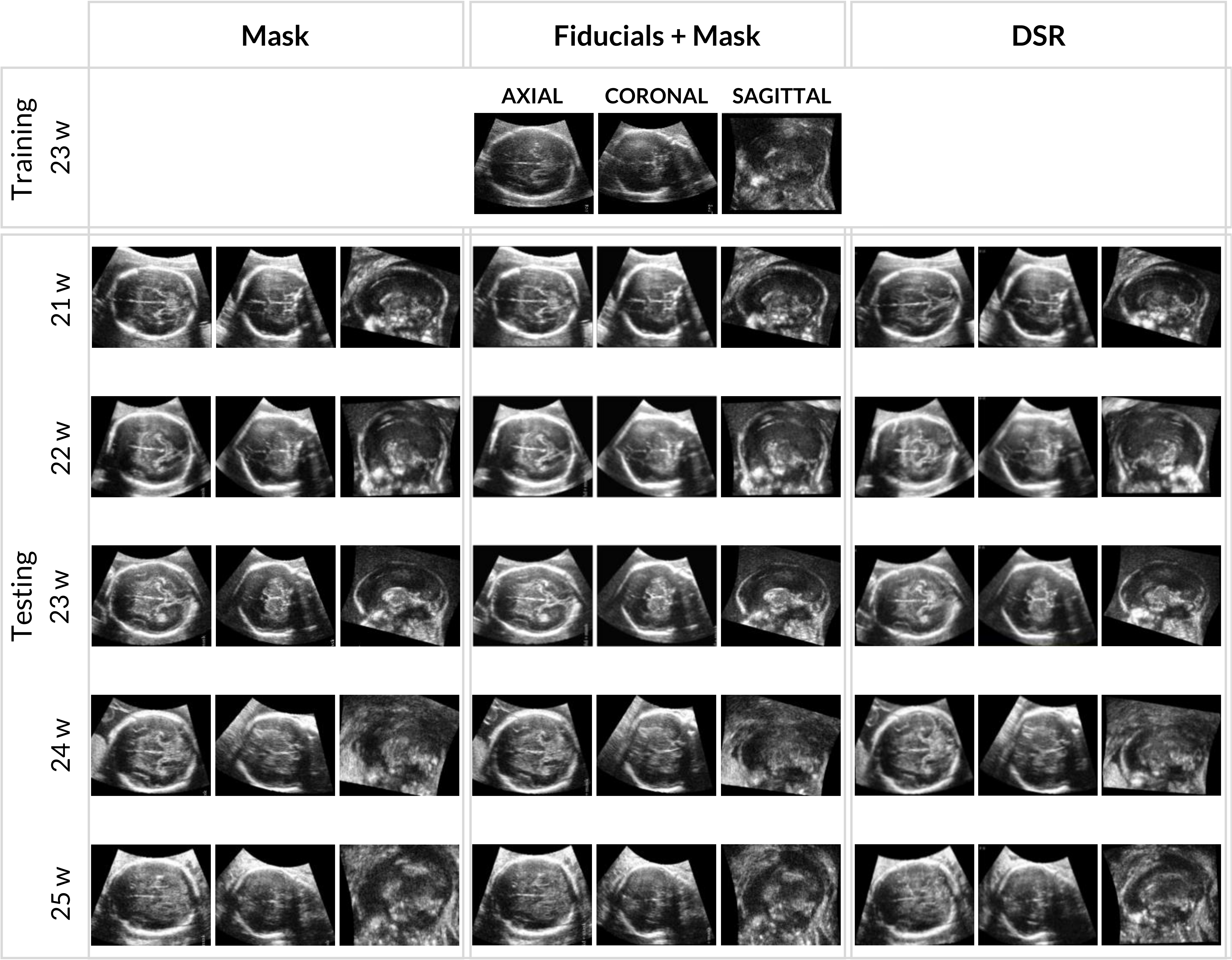}
     \caption{Axial, coronal, and sagittal views of the middle slices of each volume after registration on real fetal brain \ac{us} volumes. First column: results obtained by aligning the volumes using a mask to contour the brain used to avoid overfitting on the shape of the \ac{us} volume and then applying the automatic similarity registration (rigid registration + scale for a total of 7 degrees of freedom) provided by 3D Slicer. Second column: results obtained by aligning the volumes with the automatic registration using the fiducial points annotated by the obstetrician, followed by similarity registration using the mask that contours the brain. Third column: results obtained by aligning the volumes with \ac{dsr} registration~\cite{Zhao2022}}
     \label{fig:reg-views}
 \end{figure*}
 
\subsection{Deep Learning-based Plane Pose Regression System}
\label{subsec:methods-regression}

We base our \ac{6d} pose regression system on the network proposed in~\cite{DiVece2022} (Figure~\ref{fig:pipeline}c). We used an 18-layer residual \ac{cnn} (ResNet-18)~\cite{He2016} as a backbone for feature extraction with the pre-trained ImageNet weights~\cite{Deng2010}. We modified the network by re-initializing the fully connected layer based on the representation's dimension (nine parameters) and adding a regression head to directly output the rotation and translation representations. The network receives the \ac{us} image $I$ (128$\times$128) obtained by slicing the volume and its \ac{6d} pose with respect to the center of the fetal brain \ac{us} volume $\theta_{GT}~=~(t_x, t_y, t_z, \alpha_x, \alpha_y, \alpha_z)$. We use this information as the ground truth label for network training and validation. The \ac{cnn} learns to predict the \ac{6d} pose with respect to the same point $\theta_{Pred}=(t'_x, t'_y, t'_z, \alpha'_x, \alpha'_y, \alpha'_z)$. Specifically, the network first outputs a vector of nine parameters $\theta_{Out} = (t_{1}, t_{2}, t_{3}, r_{1}, ..., r_{6})$; the first three are used for the translation and the last six for the rotation. Then, $r_{1}, ..., r_{6}$ are used internally by our \ac{cnn} to reconstruct the rotation matrix $\bf{R}'$ in the forward pass. Differently from~\cite{DiVece2022}, we also perform image intensity augmentation. More specifically, we change brightness, contrast, and saturation. The detailed parameters of this augmentation are described in the next section.

\section{Experiments and Results}
\label{sec:exp}

In this section, we report the results of three main experiments. First, in Section~\ref{subsec:exp-reg}, we assess three different \ac{us} volume registration methods and investigate their qualitative and quantitative impact in defining a generalized inter-patient coordinate frame for the fetal brain (Section~\ref{subsubsec:exp-reg-err} and~\ref{subsubsec:exp-reg-clinical}). Besides, we illustrate the effect of these registration methods when training \ac{us} plane pose regression networks using a fetal \ac{us} volume of a single 23-week fetus to train the network and fetuses with a \ac{ga} ranging from 21 to 25 weeks for testing (Section~\ref{subsubsec:exp-reg-cnn}). In our second experiment, in Section~\ref{subsec:exp-annotations}, we investigate how consistent the manual annotations of the \ac{tv} \acp{sp} are both in terms of quality (Section~\ref{subsubsec:exp-annotations-quality}) and variability (Section~\ref{subsubsec:exp-annotations-variability}) and assess their role in evaluating the quality of volume registration and pose regression. Lastly, Section~\ref{subsec:exp-ablation} reports final pose regression results with \ac{loocv} when training data has the best registration alignment available and intensity data augmentations are performed.

\subsection{Volume Registration}
\label{subsec:exp-reg}

Before training the pose regression models, we require a set of well-aligned \ac{3d} \ac{us} volumes to generate training and validation data. Figure~\ref{fig:reg-views} reports the axial, coronal, and sagittal views of the middle slice for each \ac{us} volume after aligning them with three different volume registration methods. The first column shows the results obtained using direct similarity registration on raw \ac{us} volume data (rigid registration + scale for a total of 7 degrees of freedom) performed only on a manually annotated mask, a \ac{roi} that contours the fetal brain; this registration uses the \ac{mmi} image comparison cost metric during fitting. We provide an identity transformation for initialization. The second column shows the results obtained with a point-based registration with fiducial points annotated by an obstetrician, followed by similarity registration with a fetal brain mask. The third column shows the results obtained with a state-of-the-art intensity registration approach called \ac{dsr}. Since the method is iterative, we initialized it with the fiducial approach and performed a mono-modal rigid registration of multiple volumes~\cite{Zhao2022}.

 Table~\ref{tab:reg-scores} reports the evaluation of the different registration methods and their effects on \ac{us} plane pose regression. We performed two different types of evaluation.

\begin{table}[!t]
\centering
\caption{Evaluation of the Different Registration Methods. We Report the Root Mean Square (RMS) Error for Manually Annotated Fiducials and the RMS Error of Raw Volume Intensities, Both With Respect to a Reference 23 Week Volume. We Also Provide the Obstetrician's Assessment of Skull Alignment Based on Axial, Coronal, and Sagittal Planes on a 1 (Bad) to 5 (Good) Scale.}\label{tab:reg-scores}
\resizebox{\columnwidth}{!}{%
\begin{tabular}{cccccc} \toprule
 & \multirow{2}{*}{\begin{tabular}[c]{@{}c@{}}\bf{Evaluation}\\\bf{Metric}\end{tabular}}
 & \textbf{No registration} & \multicolumn{3}{c}{\bf{Registration}} \\ \cmidrule{4-6}
 & & & \bf{Mask} & \bf{Fid + Mask} & \textbf{\ac{dsr}}\\ \midrule
\multirow{6}{*}{21w}
 & \begin{tabular}[c]{@{}c@{}}Fiducials Errors\\(RMS [$mm$])\end{tabular}
 & 6.02 & 5.82 & \textbf{5.80} & 6.51\\ \cmidrule{3-6}
 & \begin{tabular}[c]{@{}c@{}}Volume Intensity\\ Errors (RMS)\end{tabular} 
 &\textit{62.17} & 48.63 & 47.73 & \textbf{44.04}\\ 
 \cmidrule{3-6}
 & \begin{tabular}[c]{@{}c@{}}Obstetrician\\(1-5)\end{tabular}                    
 &  & 4 & \textbf{5} & \textbf{5}\\ \midrule        
\multirow{6}{*}{22 w}
 & \begin{tabular}[c]{@{}c@{}}Fiducials Errors\\(RMS [$mm$])\end{tabular} 
 & 2.53 & 2.56 & \textbf{2.52} & 5.08\\
 \cmidrule{3-6}
& \begin{tabular}[c]{@{}c@{}}Volume Intensity\\ Errors (RMS)\end{tabular} 
& \textit{53.73} & 39.04 & 40.08 & \textbf{37.53}\\ \cmidrule{3-6}
 & \begin{tabular}[c]{@{}c@{}}Obstetrician\\(1-5)\end{tabular}                   
 & & 4 & \textbf{5} & \textbf{5}\\ 
 \midrule
\multirow{6}{*}{23 w} 
  & \begin{tabular}[c]{@{}c@{}}Fiducials Errors\\(RMS [$mm$])\end{tabular}
  & 3.34 & 3.86 & \textbf{3.32} & 6.07\\
  \cmidrule{3-6}
 & \begin{tabular}[c]{@{}c@{}}Volume Intensity\\ Errors (RMS)\end{tabular} 
 & \textit{45.27} & 38.51 & 39.09 & \textbf{37.05} \\ \cmidrule{3-6}
 & \begin{tabular}[c]{@{}c@{}}Obstetrician\\(1-5)\end{tabular}                    
 &  & 4 & \textbf{5} & \textbf{5}\\ 
\midrule 
\multirow{6}{*}{24 w} 
  & \begin{tabular}[c]{@{}c@{}}Fiducials Errors\\(RMS [$mm$])\end{tabular}
 & 3.45 & 3.74 & \textbf{3.00} & 5.36 \\
 \cmidrule{3-6}
& \begin{tabular}[c]{@{}c@{}}Volume Intensity\\ Errors (RMS)\end{tabular} 
& \textit{49.53} & 43.99 & 41.16 & \textbf{40.08} \\ \cmidrule{3-6}
 & \begin{tabular}[c]{@{}c@{}}Obstetrician\\(1-5)\end{tabular}                    
 &  & 3 & 3.5 & \textbf{\textbf{4}}\\ \midrule 
\multirow{6}{*}{25 w} 
  & \begin{tabular}[c]{@{}c@{}}Fiducials Errors\\(RMS [$mm$])\end{tabular}
 & 2.19 & 3.89 & \textbf{1.68} & 5.00 \\
 \cmidrule{3-6}
& \begin{tabular}[c]{@{}c@{}}Volume Intensity\\ Errors (RMS)\end{tabular} 
&\textit{56.14} & 45.19 & 41.43 & \textbf{34.12} \\ \cmidrule{3-6}
 & \begin{tabular}[c]{@{}c@{}}Obstetrician\\(1-5)\end{tabular}                    
 &  & 3 & 3.5 & \textbf{4}\\ 
 \midrule \midrule 
 \multirow{6}{*}{Avg} 
  & \begin{tabular}[c]{@{}c@{}}Fiducials Errors\\(RMS [$mm$])\end{tabular}
 & 3.51 & 3.97 & \textbf{3.27} & 5.60\\
 \cmidrule{3-6}
&  \begin{tabular}[c]{@{}c@{}}Volume Intensity\\ Errors (RMS)\end{tabular}
& \textit{53.37}  & 43.08 & 41.89 & \textbf{38.56} \\ \cmidrule{3-6}
 & \begin{tabular}[c]{@{}c@{}}Obstetrician\\(1-5)\end{tabular}                    
 &  & 3.6 & 4.4 & \textbf{4.6}\\ 
 \bottomrule
\end{tabular}}
\end{table}

\subsubsection{Registration error}\label{subsubsec:exp-reg-err} Registration accuracy is usually evaluated by identifying matching pairs of landmarks annotated by the clinician in the \ac{roi}. We report the error between the volumes prior to registration with a \ac{ga} ranging from 21 to 25 weeks and the one used for training (23 weeks); besides, we report the \ac{rms} error between the landmarks in $mm$ and an intensity-based \ac{rms} score for the same volumes for the three registration methods;
 
 \subsubsection{Obstetrician's evaluation}\label{subsubsec:exp-reg-clinical} We asked the obstetrician to evaluate the registration outcome for the three registration methods, shown in Figure~\ref{fig:reg-views}, by assigning a score between 1 (bad) and 5 (good), without taking into account the quality of the volumes;

 \begin{figure*}[!t]
    \centering
    \subfloat[Error distributions for the three registration methods]{\includegraphics[width=\textwidth]{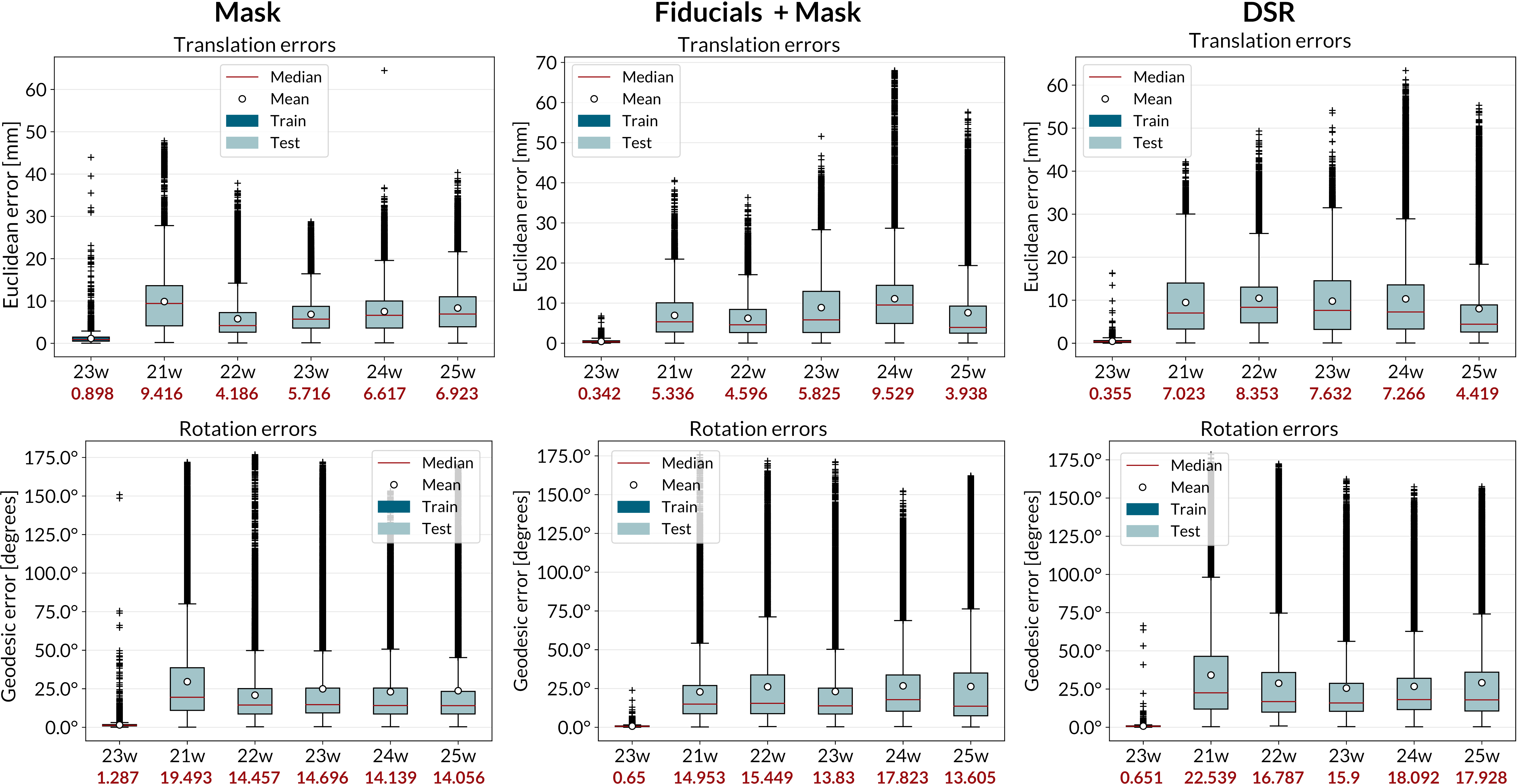}%
    \label{fig:cnn}}\\
    \subfloat[Sanity test on TV SPs - Fiducial Points + Mask]{\includegraphics[width=\textwidth]{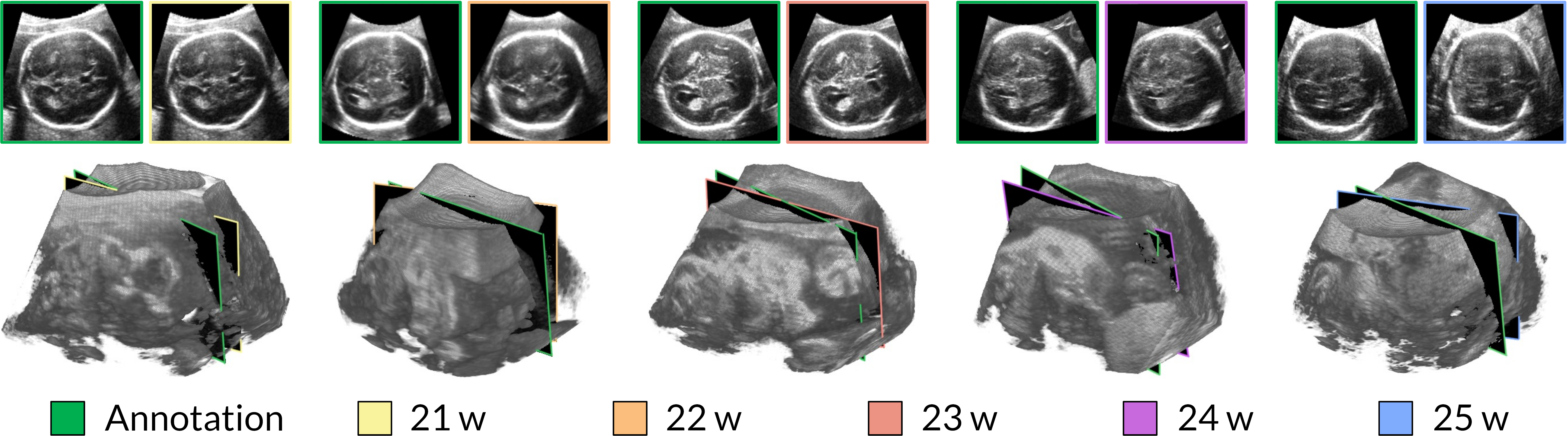}%
    \label{fig:cnn-sc}}
    \caption{CNN results for the different registration methods, trained on a single 23w volume. (a) Translation and rotation error distributions for both planes acquired at random coordinates and planes acquired around the annotated \ac{tv} \ac{sp} to analyze the generalization capabilities of the network with the three registration methods. (b) TV SP prediction performed by the regression CNN. The green and the colored boxes indicate the ground truths and the predictions, respectively. An obstetrician within the Unity environment manually annotated the ground truth poses of the TV SPs}
    \label{fig:cnn-res}
\end{figure*}

\subsection{Registration impact on CNN results}\label{subsubsec:exp-reg-cnn}
We measure the errors of our pose regression \ac{cnn} according to the different ``ground truths'' resulting from different registration methods. We evaluate to which extent the different volume registration methods affect the perceived \ac{us} plane pose regression results.
  \subsubsection*{Implementation Details} Our framework is implemented in PyTorch and trained using a single Tesla\textregistered{} A100-SXM4-40GB hosted on the Computer Science network at University College London. The network was trained for 50 epochs with a batch size of $K$ = 64 using Adam optimizer, with a learning rate of 0.0001 and exponential decay rates $\beta_1$ and $\beta_2$ of 0.9 and 0.999, respectively. We choose the best model weights considering \ac{mse} obtained on the validation set (20\% of the training set). 
 \subsubsection*{Experiments} The network was trained on phantom data and fine-tuned on real ones~\cite{DiVece2022}. Specifically, we fine-tuned the network on planes extracted from a fetal brain \ac{us} volume with a \ac{ga} of 23 weeks ($f_{1}$, 22029 images). We tested it on planes from five volumes obtained with a single acquisition of different fetuses ($f_2,..., f_6$) ranging from a \ac{ga} of 21 to 25 weeks to understand how well the model generalizes over different shapes and sizes. Images were resized to 128$\times$128, preserving the same aspect ratio, and cropped and centered to avoid visible sharp edges that could cause overfitting. We augmented the training set by randomly changing the images' brightness, contrast, and saturation to a value between 0 and 1. To this aim, we used the \texttt{torchvision.transforms.ColorJitter} class that is available in Pytorch for transforming and augmenting images. We employed the Euclidean distance between the two planes to evaluate the translation results, reported in $mm$. For rotation, we display errors as the geodesic distance to ground truth in $degrees$, more suitable for the geometric interpretation of the distance between two \ac{3d} rotations and defined as $Error_{Rotation} = \arccos ((\bf{R}''_{00}+\bf{R}''_{11}+\bf{R}''_{22}-1)/2)$, where $\bf{R}''~=~\bf{R}'^{-1}$. Table~\ref{tab:reg-scores} reports the median errors for translation and rotation obtained on the testing volumes for the three registration methods. Figure~\ref{fig:cnn-res} reports the translation and rotation error distributions for fetal brain \ac{us} volumes ranging from a \ac{ga} of 21 weeks to 25 weeks to analyze the generalization capability of the network in the three registration methods. Besides, we performed a sanity test using the manually annotated \ac{tv} \acp{sp} for the registration using the fiducial points and the mask. The sectional images were saved and fed into the network to estimate their pose. We plotted the two planes within the volume in Unity to visually evaluate the distance between the annotated \ac{tv} \acp{sp} and the predicted ones. The predicted planes were also fed into a pre-trained\footnote{\href{ttps://github.com/baumgach/SonoNet-weights}{https://github.com/baumgach/SonoNet-weights}} SonoNet, a \ac{cnn} that can automatically detect 13 fetal standard views in freehand \ac{2d} \ac{us} data~\cite{Baumgartner2017a}, in its Pytorch implementation. Figure~\ref{fig:cnn-res} reports the annotated \ac{tv} \acp{sp} (green edges) and the ones having the pose predicted by the regression network in their sectional view and within the volumes.
\subsection{Standard Planes Annotations}
\label{subsec:exp-annotations}

An obstetrician annotated the \acp{sp} for all the \ac{3d} \ac{us} fetal brain volumes previously registered using the Unity simulator detailed above.

\subsubsection{Quality of Annotations}\label{subsubsec:exp-annotations-quality}
\begin{figure}[!t]
     \centering
     \includegraphics[width=\columnwidth]{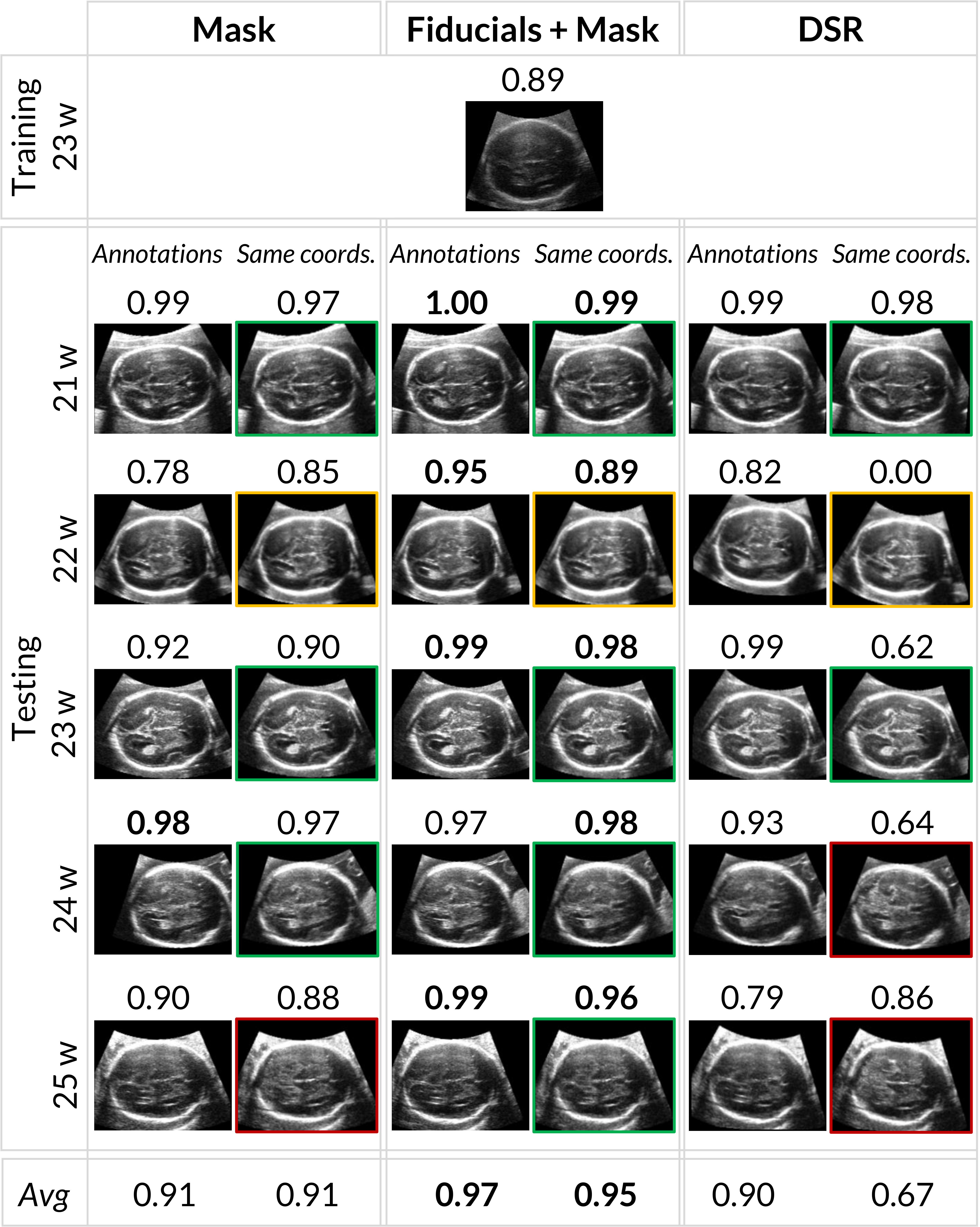}
    \caption{Evaluation of different registration methods according to \ac{sp} alignment. For each method, the first column denotes manually annotated \acp{sp} for each volume, and the second denotes a slice at fixed coordinates where the \ac{sp} is expected, assuming the same coordinates as the reference volume (23 w). Colors denote whether the \ac{sp} follows clinical guidelines~\cite{Salomon2011}. Green box: the \ac{sp} is of good quality; Orange box: the \ac{sp} does not contain the \ac{csp}, which should be visible; Red box: the \ac{sp} contains the cerebellum, which should not be visible. The numbers denote SonoNet confidence in correctly identifying the correct \ac{sp}}
     \label{fig:sps}
 \end{figure}

To evaluate the quality of the annotated \ac{tv} \acp{sp}, we use SonoNet in its PyTorch implementation. We report the annotated \ac{tv} \acp{sp} for the various volumes and the registration methods in Figure~\ref{fig:sps}. SonoNet was able to classify all the annotated \acp{sp} as \ac{tv} \acp{sp}, the brain view at the posterior horn of the ventricle. We then applied the coordinates of the \ac{tv} \ac{sp} annotated on the training volume (23 w) to the other planes. The synthetic images obtained for the different volumes were fed again into SonoNet to understand if the network could still recognize the planes as standard views. All the planes were recognized as \ac{tv} \acp{sp}, except for the volume with a \ac{ga} of 25 weeks obtained using the automatic registration. The confidence of the classification (value between 0 and 1) is reported on top of each image in Figure~\ref{fig:sps}. For each registration approach, the first column shows the \ac{tv} \acp{sp} obtained from the annotations. In contrast, the second column shows the \ac{tv} \acp{sp} obtained by using the coordinates of the \ac{tv} \ac{sp} annotated on the training volume. An obstetrician evaluated the appearance of \acp{sp} shown in the second column to make sure that the obtained planes follow the relevant clinical guidelines~\cite{Salomon2011}. In this context, some of the planes are of poor quality due to the absence of the \ac{csp}, shown in Figure~\ref{fig:sp}, and are marked in orange; other planes display the cerebellum which should not be visible and are marked in red. At the bottom, we report the average score, including the training volume (23 w).

\subsubsection{Variability in SPs Annotations}
\label{subsubsec:exp-annotations-variability}
\begin{figure*}[!t]
    \centering
    \includegraphics[width=\textwidth]{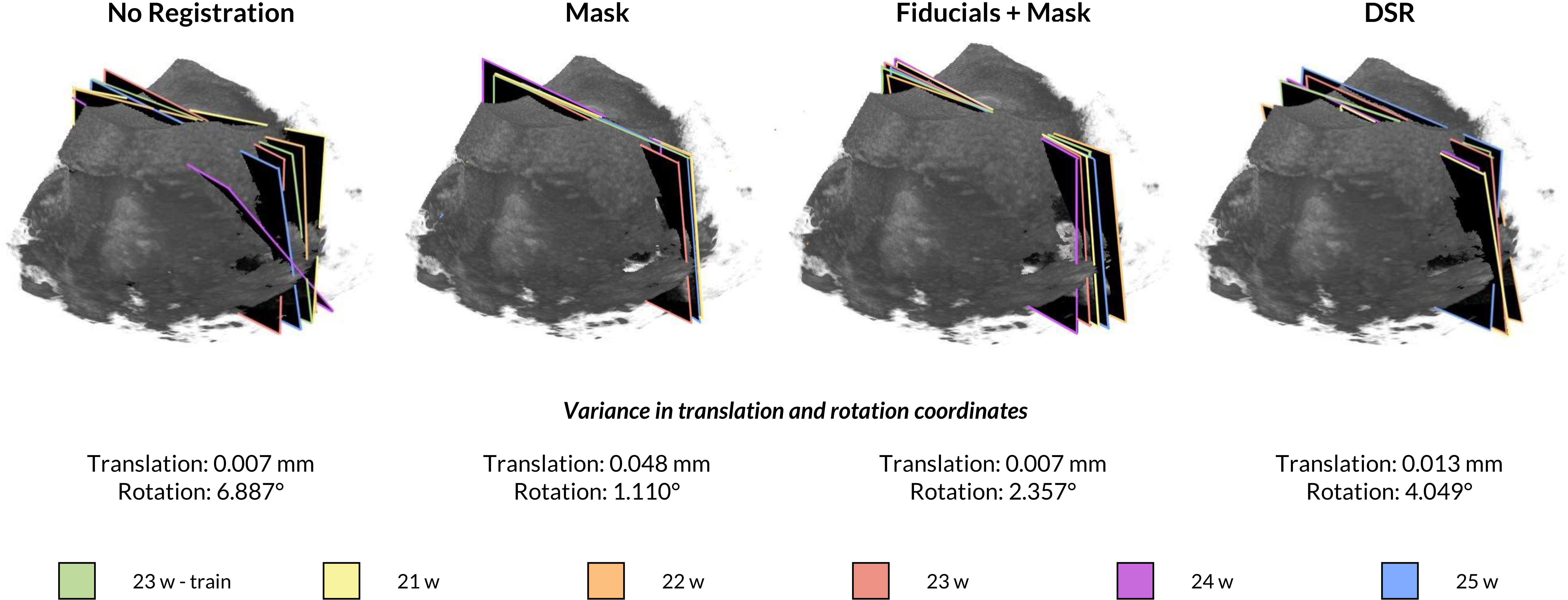}
    \caption{TV SPs annotated by the obstetrician for the three registration methods and variance in translation and rotation coordinates}
    \label{fig:annotations}
\end{figure*}

We evaluate the variability in annotations by reporting the standard deviation of the poses of the \ac{tv} \acp{sp} annotated by the obstetrician in the various volumes. The internal variance of the entire set provides the quality of the data. We report the variance in translation and rotation of the annotated \ac{tv} \acp{sp} for each registration approach. For translation, we computed the coordinates of the centroid $c_{transl} = x_{c},y_{c},z_{c}$ for each group made of the training volume and the five testing volumes ($i = 1,...,6$) as the mean on of three coordinates:
\begin{equation}
  x_c = \frac{\sum_{i=1}^{6}x_i}{6}, y_c = \frac{\sum_{i=1}^{6}y_i}{6}, z_c = \frac{\sum_{i=1}^{6}z_i}{6}  
\end{equation}
Then, we computed the Euclidean distance between each \ac{tv} \ac{sp} and the centroid:
\begin{equation}
    d_{i,transl} = \sqrt{(x_i-x_c)^2+(y_i-y_c)^2+(z_i-z_c)^2}
\end{equation}
Lastly, we computed the \ac{rms} of these distances:
\begin{equation}
    RMS_{transl} = \sqrt{\frac{1}{6} \sum_{i=1}^{6}d_{i,transl}^2}
\end{equation}
We computed the Chordal L2-averaging of the rotation matrices for rotation, following the approach presented in~\cite{Hartley2021}. This is achieved by finding the rotation $R_c$ that minimizes the cost $\sum_{(i,j)\in N}^{}  \left \| R_{ij}R_i - R_j \right \|_F^2$. The above model can be solved without enforcing the orthogonality constraint as a least squares problem through vectorization and singular value decomposition. After that, all the orthogonal constraints are enforced by finding the nearest orthogonal matrices through polar decomposition.
Then, we computed the geodesic distance between the rotation matrix of each \ac{tv} \ac{sp} and the average rotation matrix (angle of residual rotation):
\begin{equation}
    d_{i,rot}(R_i, R_c) = d_{i,rot}(R_iR_c^T, I) = \left \| log(R_i R_c ^T) \right \|_2
\end{equation}
where the norm is the Euclidean norm in $\mathbb{R}^3$. The angular distance function $d_{i,rot}(R_i, R_c)$ is equal to the rotation angle $\angle (R_iR_c^T)$. Starting from the quaternion representations, it is possible to easily compute the angular distance between two rotations. If $r_i$ and $r_c$ are quaternion representations of $R_i$ and $R_c$ respectively, and $\theta = d_{i,rot}(R_i, R_c)$, then $\theta = 2 arccos(|s|)$, where $(s,\bf{v}) = r_c^{-1} \cdot r_i$. The absolute value sign in $s$ is required to account for the sign ambiguity in the quaternion representation of the rotation $R_i^T R_c$. The positive sign is chosen so that the angle $\theta$ lies in the range $0 \leq \theta \leq \pi$, as required. Hence, the distance $d_{i,rot}(R_i, R_c)$ is equal to the angle $\theta$ belonging to the rotation $R_i R_c^T$. As before, we computed the root mean square of the distances:
\begin{equation}
    RMS_{rot} = \sqrt{\frac{1}{6} \sum_{i=1}^{6}d_{i,rot}(R_i, R_c)^2}
\end{equation}
The results are reported in Figure~\ref{fig:annotations} along with the appearance of the \ac{tv} \acp{sp} annotated by the obstetrician for the three registration methods.

\subsection{Ablation Study on the Pose Regression CNN} 
\label{subsec:exp-ablation}

We performed an ablation study on the volumes registered using the combination of fiducial points and masks. First, we present the results for the \ac{us} plane pose regression with and without data augmentation on the training set. Then, we extend the training set and combine data augmentation with \ac{loocv}.

\subsubsection{Data augmentation}
\label{subsec:exp-ablation-da}
Data augmentation artificially boosts the size and variance of the training dataset by including transformed copies of the training examples. This is especially useful in medical imaging, where data augmentation is applied to expand training data, address the class imbalance and improve model generalization. To understand to which extent data augmentation could benefit our training and increase the generalization over different shapes and sizes, we augmented the training set, as detailed above.
\subsubsection{Leave-one-out Cross-Validation}\label{subsec:exp-ablation-loocv}
\begin{figure*}[!t]
\centering
\subfloat[Exp 1 - Test: 23T w]{\includegraphics[width=\columnwidth]{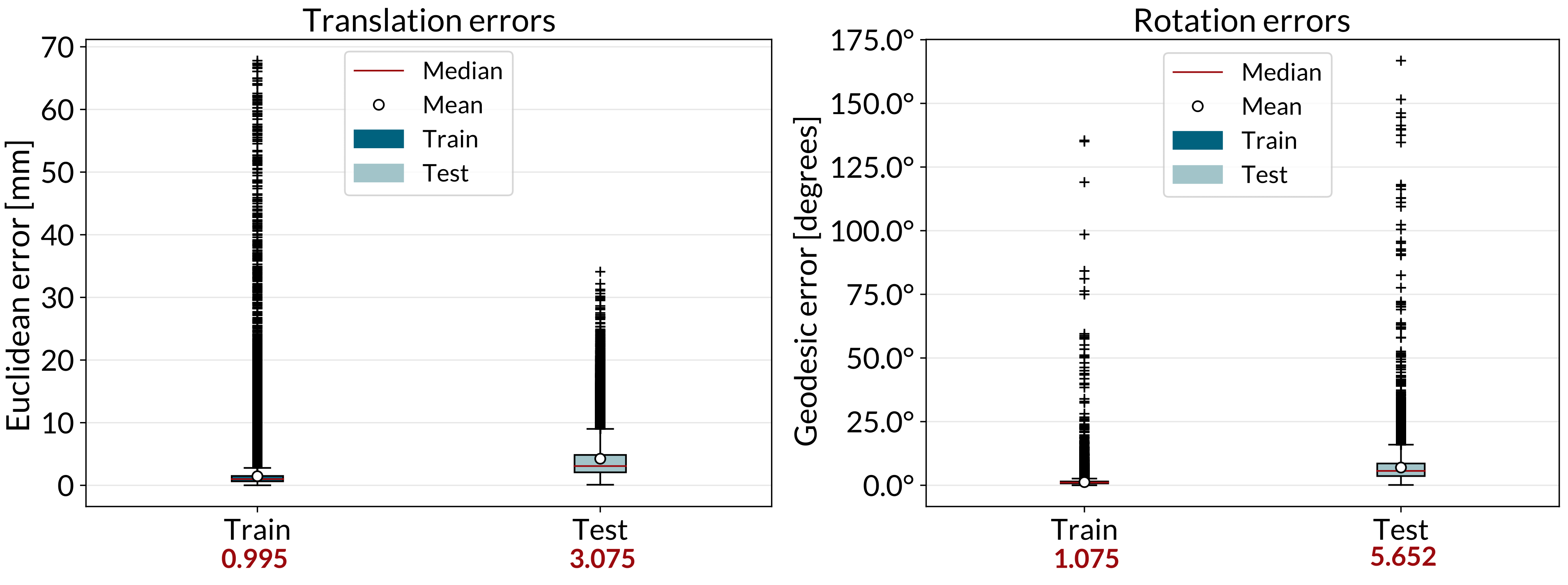}%
\label{subfig:cnn-loocv1}}
\hfil
\subfloat[Exp 4 - Test: 23 w]{\includegraphics[width=\columnwidth]{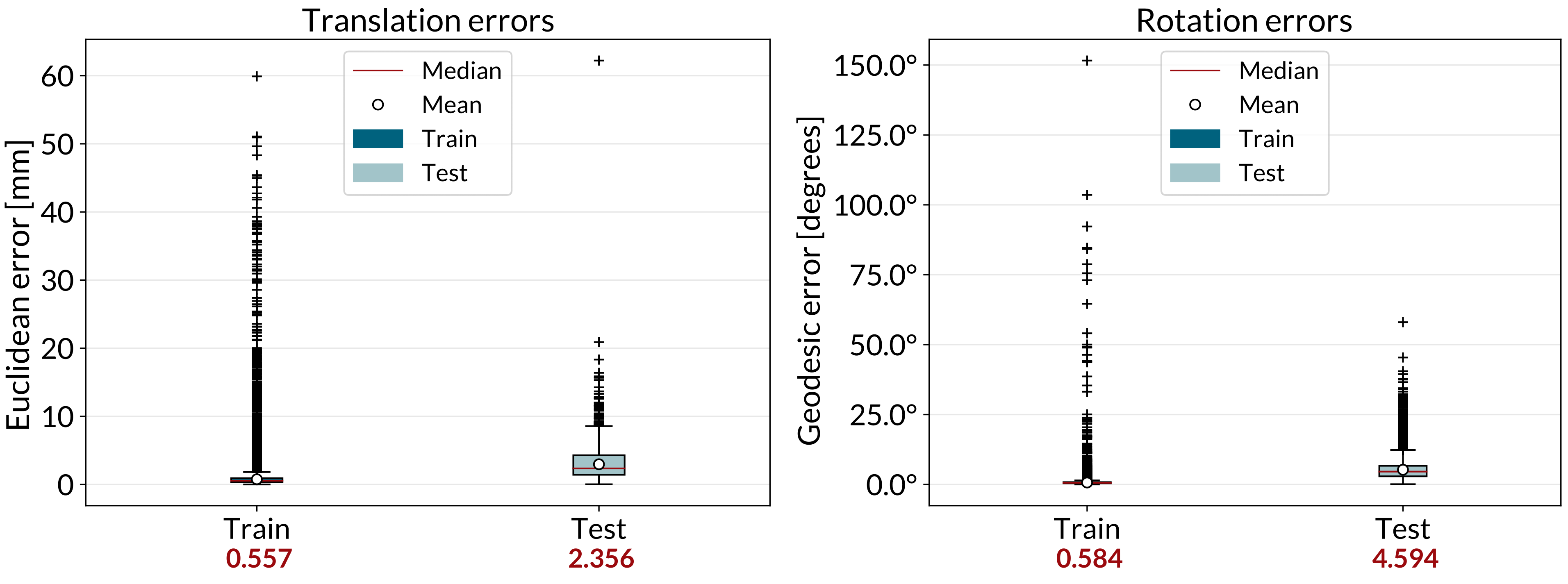}%
\label{subfig:cnn-loocv4}}\\
\subfloat[Exp 2 - Test: 21 w]{\includegraphics[width=\columnwidth]{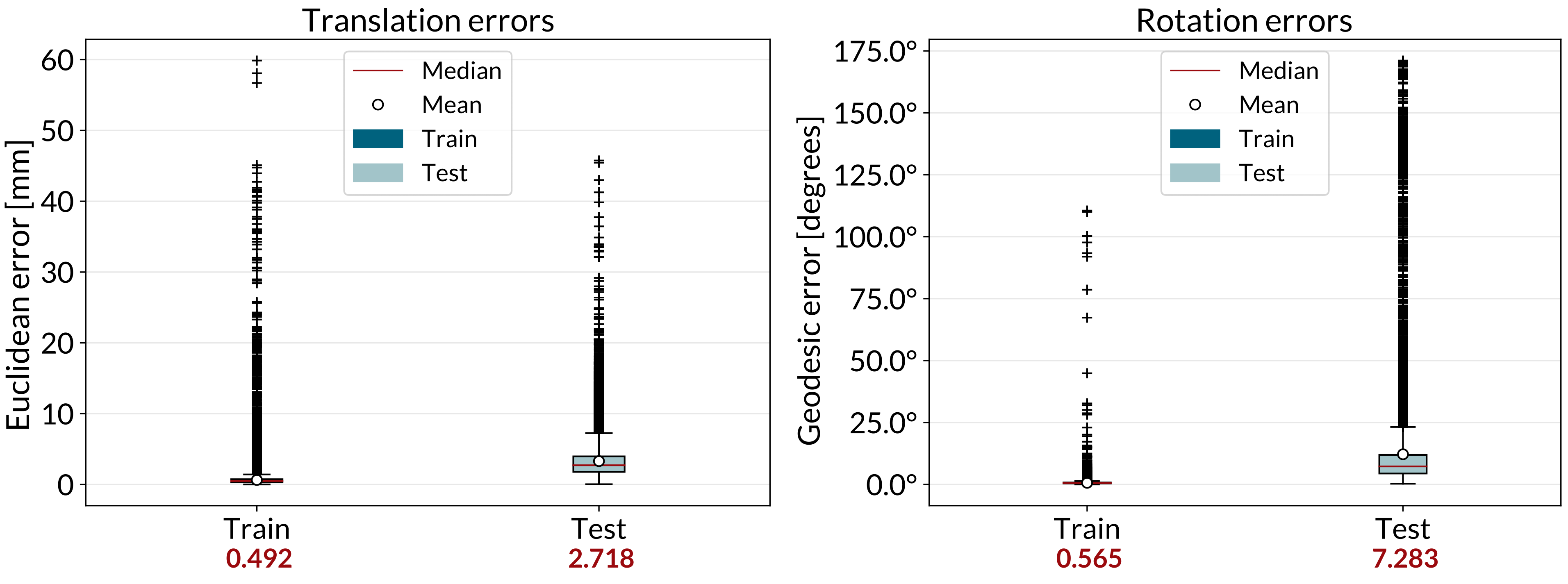}%
\label{subfig:cnn-loocv2}}
\hfil
\subfloat[Exp 5 - Test: 24 w]{\includegraphics[width=\columnwidth]{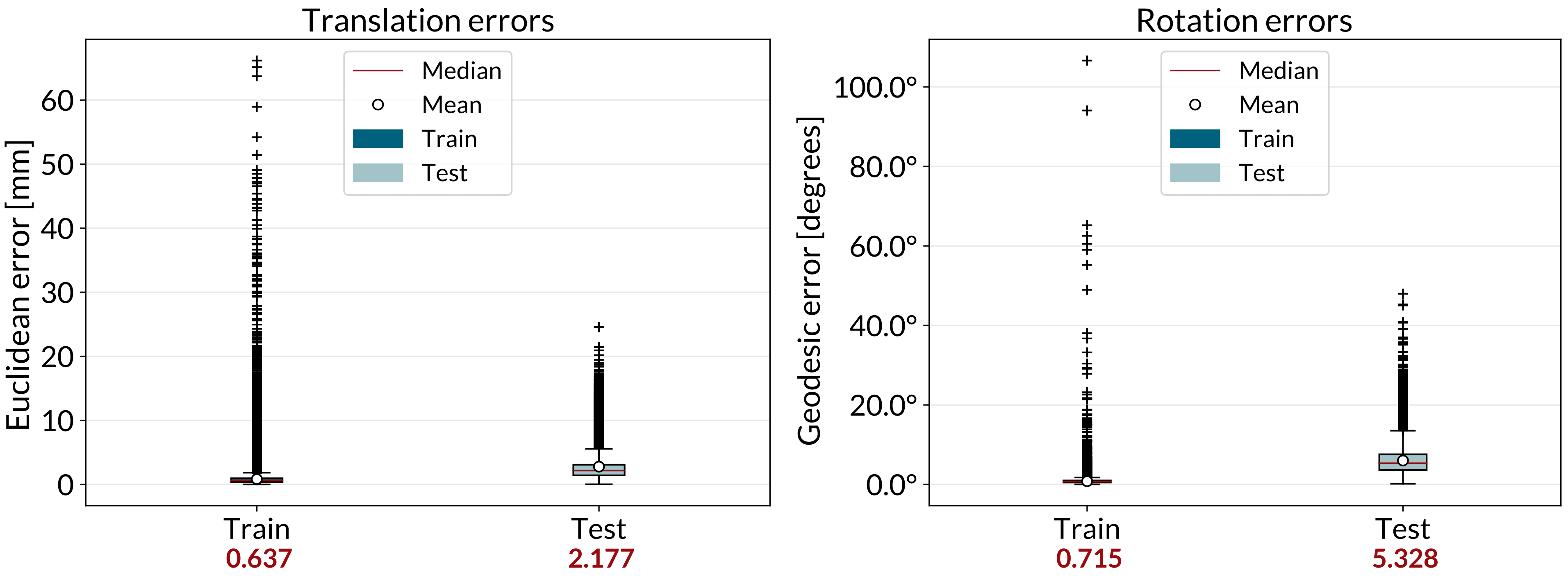}%
\label{subfig:cnn-loocv5}}\\
\subfloat[Exp 3 - Test: 22 w]{\includegraphics[width=\columnwidth]{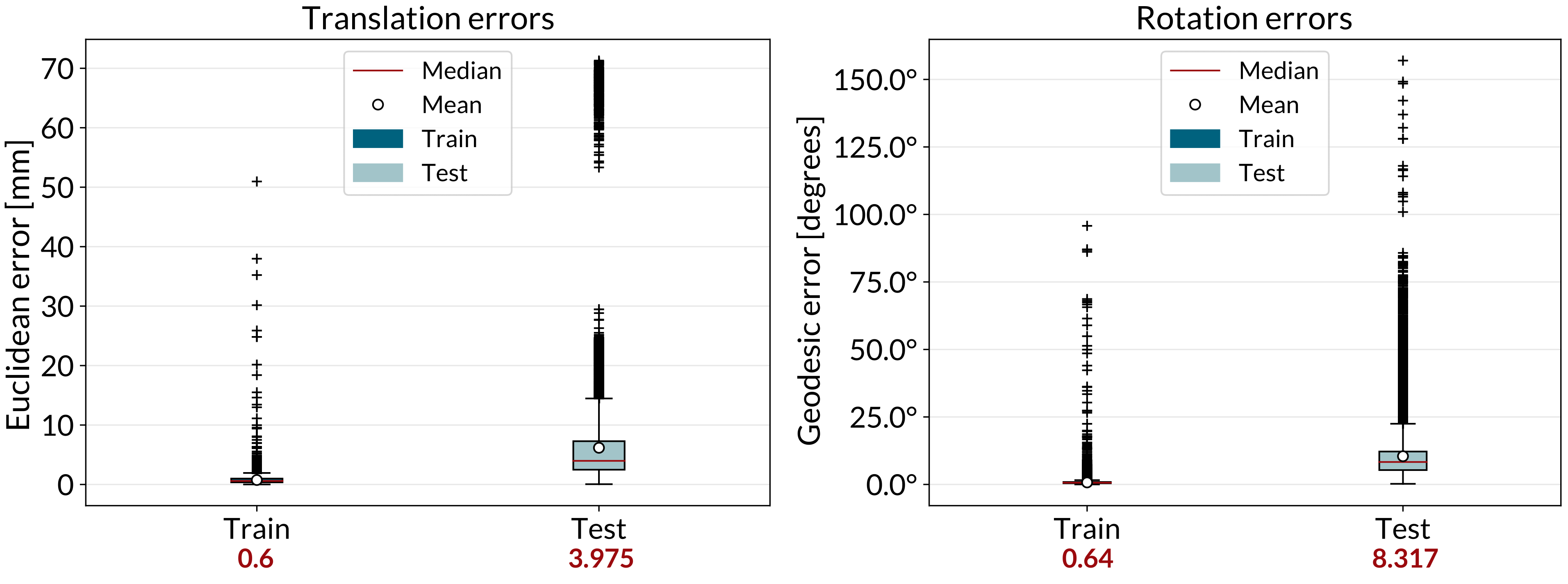}%
\label{subfig:cnn-loocv3}}
\hfil
\subfloat[Exp 6 - Test: 25 w]{\includegraphics[width=\columnwidth]{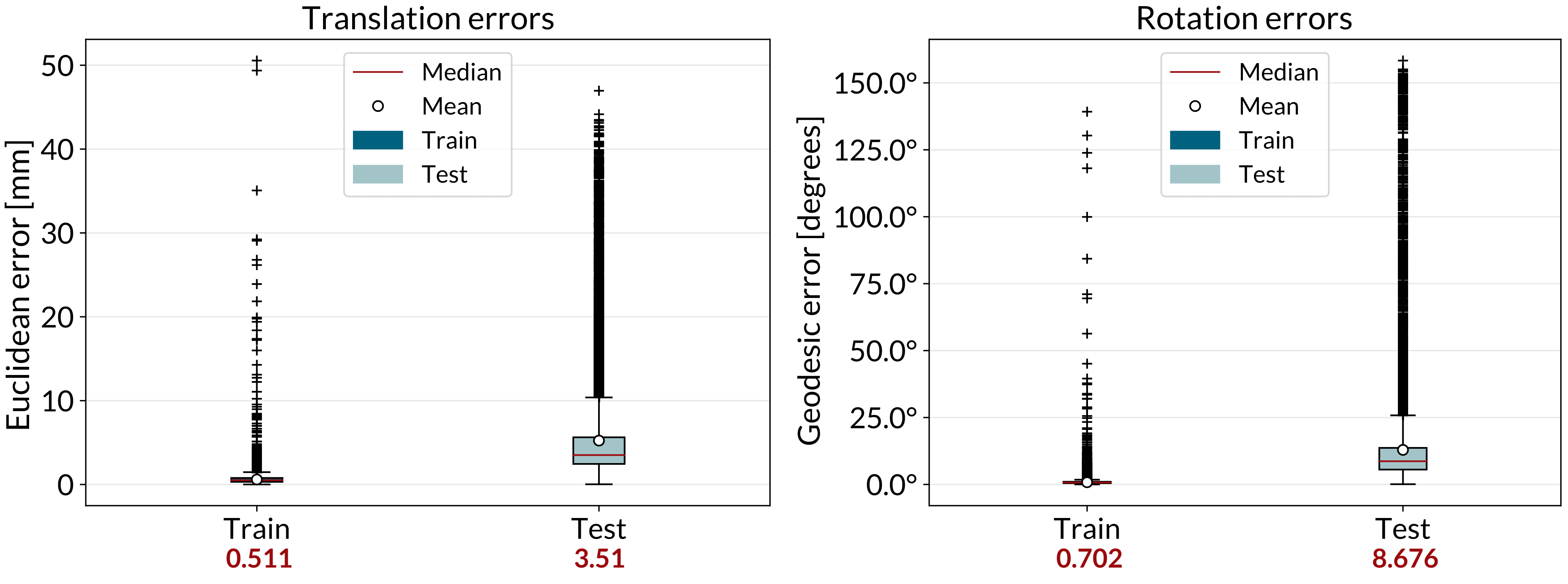}%
\label{subfig:cnn-loocv6}}\\
\subfloat[Sanity test on TV SPs with LOOCV - Fiducial Points + Mask]{\includegraphics[width=1\textwidth]{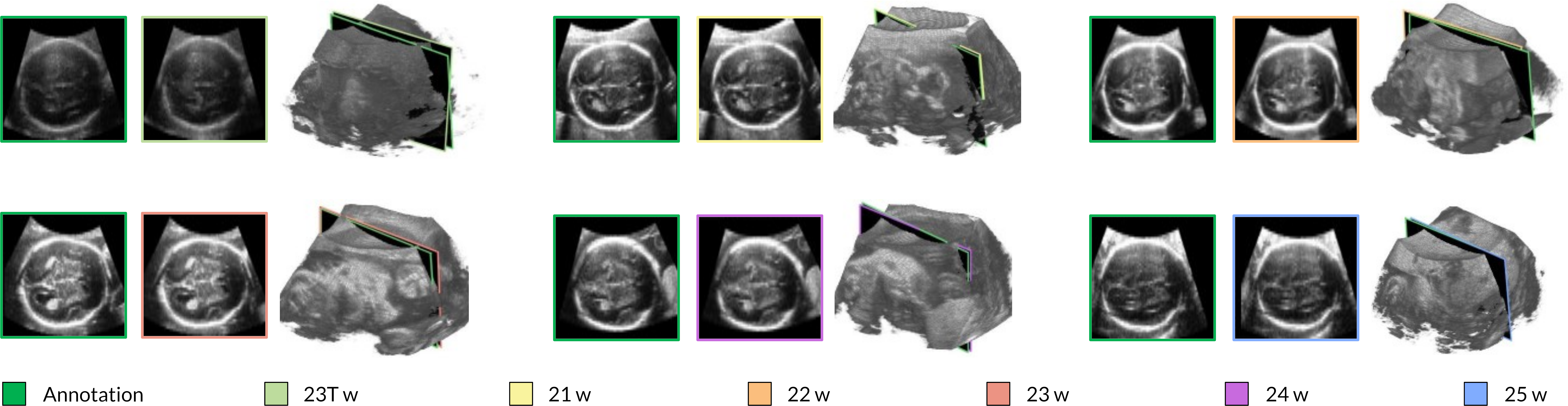}\label{subfig:cnn-loocv-sc}}
\caption{Results obtained for the experiments performed with the LOOCV. (a-f) Translation and rotation error distributions for the LOOCV experiments for the six cases. (g) TV SP prediction performed by the regression CNNs. The green and the colored boxes indicate the ground truths and the predictions, respectively. An obstetrician within the Unity environment manually annotated the ground truth poses of the TV SPs. 23T indicates the volume having a GA of 23 weeks used for reference in the registration and training in the previous experiments. The consistency between the annotated and predicted \acp{sp} is demonstrated by the fact that the planes show the same anatomical structures.}
\label{fig:cnn-loocv}
\end{figure*}

The localization errors increase when the training distribution differs from the testing one. A better training and testing data distribution design can accurately reflect the model's performance during application. Hence, to estimate the performance of our algorithm in making predictions on data not used to train the model, we performed \ac{loocv} experiments using the volumes with a \ac{ga} ranging from 21 to 25 weeks, including the one originally used for training (23 weeks), for a total of six volumes ($N=6$). \ac{loocv} is a special case of $k$-fold cross-validation with $k=N$, the number of volumes. \ac{loocv} involves one fold per volume \emph{i.e.}, each volume plays the role of the validation set. The $(N-1)$ volumes play the role of the training set. With least-squares linear, a single model performance cost is the same as a single model. The average error on the test set is calculated by fitting on the volumes not used in training and gives us an idea of how well the model will perform on data it has not previously seen. We then calculate the error on the test set $Test~Err_{avg}$ to be the average of all the errors on the six test sets $Test~Err_i$:
\begin{equation}
    Test~Err_{avg} = \frac{1}{N}\sum_{i=1}^{N}~Test~Err_i
    \label{eq:loocv}
\end{equation}
Figures~\ref{fig:cnn-loocv}a-f show the translation and rotation error distributions for the \ac{loocv} experiments for the different trained models. Figure~\ref{fig:cnn-loocv}g reports the results for the sanity test performed using the manually annotated \ac{tv} \acp{sp}, as previously detailed in Section~\ref{subsubsec:exp-reg-cnn}). Table~\ref{tab:cnn-loocv-res} reports the median, mean~$\pm$~standard~deviation, maximum, and minimum errors and the average error with and without data augmentation (Equation~\ref{eq:loocv}).
\begin{table*}[!t]
\centering
\caption{Translation and Rotation Errors of Our Method For the LOOCV Experiments on Including Data Augmentation. Norm: Euclidean Distance, GE: Geodesic Error, DA: Data Augmentation, SD: Standard Deviation, 23T indicates the volume having a GA of 23 weeks used for reference in the registration and training in the previous experiments}
\label{tab:cnn-loocv-res}
\resizebox{\textwidth}{!}{%
\begin{tabular}{cccccccccc} \toprule
\multirow{2}{*}{\bf{Test Volume}} & \multirow{2}{*}{\bf{DA}} & \multicolumn{4}{c}{\bf{Translation - Norm [mm]}} & \multicolumn{4}{c}{\bf{Rotation - GE [deg]}} \\ \cmidrule{3-10}
& Yes & \textit{Median} & \textit{Mean$\pm$SD} & \textit{Min} & \textit{Max} & \textit{Median} & \textit{Mean$\pm$SD} & \textit{Min} & \textit{Max} \\ \midrule
\multirow{1}{*}{23T w} 
& Yes & 3.07 & 4.24$\pm$3.71 & 0.09 & 34.09 & 5.65 & 6.97$\pm$6.41 & 0.13 & 166.77\\ \midrule
\multirow{1}{*}{21 w} 
& Yes & 2.72 & 3.30$\pm$2.67 & 0.04 & 45.75 & 7.28 & 12.18$\pm$19.03 & 0.29 & 171.10 \\ \midrule
\multirow{1}{*}{22 w} 
& Yes & 2.18 & 2.79$\pm$2.36 & 0.03 & 24.61 & 5.33 & 6.02$\pm$3.65 & 0.18 & 47.97 \\ \midrule
\multirow{1}{*}{23 w} 
& Yes & 2.36 & 2.96$\pm$2.02 & 0.03 & 62.22 & 4.59 & 5.27$\pm$3.48 & 0.09 & 58.03 \\ \midrule
\multirow{1}{*}{24 w} 
& Yes & 3.97 & 6.17$\pm$8.09 & 0.05 & 71.29 & 8.31 & 10.45$\pm$9.48 & 0.23 & 156.98 \\\midrule
\multirow{1}{*}{25 w} 
& Yes & 3.51 & 5.23$\pm$5.40 & 0.03 & 46.96 & 8.67 & 12.91$\pm$16.27 & 0.09 & 158.32 \\\midrule
\multirow{2}{*}{$Test~Err_{avg}$} 
& No & 3.85 & 4.96$\pm$3.98 & 0.06 & 42.12 & 7.44 & 7.05$\pm$8.23 & 0.19 & 120.41 \\ \cmidrule{2-10}
& Yes & 2.97 & 4.11$\pm$4.04 & 0.05 & 47.48 & 6.63 & 8.96$\pm$9.72 & 0.16 & 126.52
\\\bottomrule
\end{tabular}%
}
\end{table*}

\section{Discussion}
\label{sec:discussion}
First, we note that even though all the utilized volumes in this study were acquired with the same protocol to capture the \ac{tv} \ac{sp} plane, this does not mean the anatomies in different volumes are well aligned. This is evident from pre/post-registration results in Table~\ref{tab:reg-scores} and Figure~\ref{fig:annotations}. The reasons can include different fetus positions in the womb, different fetus brain sizes, operator variability, and \ac{us} machine settings.

Our volume registration results in Table~\ref{tab:reg-scores} indicate that Fiducials+Mask provides the best alignment in terms of fiducial errors. We argue that the alignment of such anatomical structures is of utmost importance in our application context. The results in Figure~\ref{fig:sps} further support that Fiducials+Mask provides the best alignment of clinically relevant anatomy by assessing the expected location of the \ac{tv} \ac{sp} plane across different fetuses, showing higher consistency than other methods. On the other hand, \ac{dsr} is the best performing in terms of intensity errors (Table~\ref{tab:reg-scores}). However, this may indicate overfitting since the same anatomical landmarks in different scans of different fetuses do not necessarily have the same intensity values. Regarding the obstetrician evaluation, \ac{dsr} slightly outperforms Fiducials+Mask. This metric reflects a qualitative skull alignment in axial/coronal/sagittal brain views. Finally, the results from Figure~\ref{fig:annotations} suggest that annotated \ac{tv} \acp{sp} on each fetus are the most consistent in translation for Fiducials+Mask, and the most consistent in rotation for Mask. While rotation and translation alignment favor different methods, we observe that the best translation alignment of \ac{tv} \acp{sp} (Fiducials+Mask) overlaps with the best alignment of anatomical features overall. In summary, the metrics most directly linked to the correct alignment of structures inside the brain (fiducial errors and \ac{tv} \ac{sp} image assessment) suggest that Fiducials+Mask provides the best registration results. With this in mind, we believe this method provides the most appropriate registration to validate the generalization of brain structure locations across different fetuses and use it to train and test our \ac{cnn}. After this alignment, a set of \ac{sp} from different fetuses, manually annotated by an obstetrician, has a variance of 0.007 $mm$ and 2.357\degree~in translation and rotation, respectively. These values are indicative of the ground truth uncertainty and therefore provide an estimated lower bound for our \ac{cnn} pose regression accuracy.

The results in Figure~\ref{fig:cnn} demonstrate that different volume registrations significantly impact the interpretation of regression \ac{cnn} results, simultaneously affecting the trained model's perceived quality and the ground truth. While this effect was mentioned in~\cite{DiVece2022} as a study limitation, here we quantify its impact. We note that better \ac{cnn} results do not necessarily mean better registration quality due to variations in the ground truth. We present these results to highlight that \ac{cnn} results alone are insufficient to assess registration quality and only indicate how well the training and test data fit together.

After establishing Fiducials+Mask as the most trustworthy ground truth alignment, we fully assess our \ac{cnn} with our \ac{loocv} study (Figure~\ref{fig:cnn-loocv}). Table~\ref{tab:cnn-loocv-res} shows that we outperform the pose regression results obtained in~\cite{DiVece2022} due to many factors. These include the more rigorous volume registration process, larger training sets that include various \acp{ga}, additional image intensity data augmentation, and removal of a fully connected layer before final regression. The proposed \ac{cnn} successfully generalizes pose regression to an unseen fetal brain. Specifically, data augmentation decreased median errors by 22.86\% and 10.88\% for translation and rotation, respectively. Besides, by extending the training sets, we could further decrease the median errors and obtain a better generalization. Our model is designed to be size invariant by performing pose regression in normalized coordinates with respect to the brain limits. However, \ac{ga} does not only affect size but also shape. Indeed, including different \acp{ga} covers a wide range of shapes and sizes, enabling us to understand our model's current generalizability and limitations.

Despite our promising results relative to the existing literature \cite{DiVece2022,Yeung2021,Yeung2022}, our regression network still produces outlier predictions with large errors. This is to be expected, given that we perform estimations based on single \ac{us} scans, which can be noisy. Our median results still suggest that the large majority of predictions has a relatively small error; hence, this is a promising backbone for any future work aiming at pose regression from continuous \ac{us} video where sparse erroneous predictions could be filtered with temporal models (\emph{e.g.}, sliding window filtering/regularization, LSTMs, transformers).

\section{Conclusion}
\label{sec:conlusion}

In the context of fetal \ac{us} plane pose regression, our study highlights the need for an application-specific registration methodology to align training and test \ac{us} volumes from different fetuses. The algorithm design and its evaluation should focus on the explicit alignment of anatomical features rather than volume intensities.

The obtained registration results provide promising evidence that assuming generalized coordinates for the fetal brain is a valid assumption within a small tolerance, especially in the context of anatomies present in the \ac{tv} \ac{sp}. However, our data only includes 6 volumes, and further analysis will require detailed annotation of additional volumes. Besides, in our experiments, we make use of inter-patient volume-to-volume \ac{us} registration, a domain with limited tailored algorithms. Intensity-based approaches such as \ac{dsr} or accumulated pairwise estimates (APE)~\cite{Wachinger2013} are not optimal for inter-patient registration. Potential improvements could be achieved with landmark-based registration approaches~\cite{gomez2017fast,wright2019complete,deepreg2020}. However, these are not off-the-shelf applicable to our inter-fetus registration problem; thus, adapting them requires further work and additional training data.

We fully re-assess the results of our \ac{cnn} pose regression network for localizing fetal brain \ac{us} scans in light of the improved registration methodology. We can estimate plane poses within the brain with a median translation error of 2.97 $mm$ and a median rotation error of 6.63\degree. These results are promising, given that we are localizing images of a previously unseen fetus from a single frame. However, there are still outlier predictions with large errors (refer to maximum errors in Table~\ref{tab:cnn-loocv-res}). We believe the key to address this challenge is to move away from single-frame estimation and take temporal cues into account.

This work could potentially be generalized to other anatomical regions of the fetus, such as the abdomen; however, the definition of a generalized reference frame would still be challenging due to increased deformations. 

We will also assess the potential of the current work toward active guidance of sonographers during \acp{sp} acquisition for fetal biometry. This could include automated feedback signals to guide a novice from an arbitrary \ac{us} plane toward the target \ac{sp}. 

\bibliography{references}
\bibliographystyle{IEEEtran}


\end{document}